# The inverted Pendulum: A fundamental Benchmark in Control Theory and Robotics

Olfa Boubaker
National Institute of Applied Sciences and Technology
INSAT, Centre Urbain Nord BP. 676 – 1080 Tunis Cedex, Tunisia
olfa.boubaker@insat.rnu.tn

*Abstract*— **For at least fifty years, the inverted pendulum has been the most popular benchmark, among others, for teaching and researches in control theory and robotics. This paper presents the key motivations for the use of that system and explains, in details, the main reflections on how the inverted pendulum benchmark gives an effective and efficient application. Several real experiences, virtual models and web-based remote control laboratories will be presented with emphasis on the practical design implementation of this system. A bibliographical survey of different design control approaches and trendy robotic problems will be presented through applications to the inverted pendulum system. In total, 150 references in the open literature, dating back to 1960, are compiled to provide an overall picture of historical, current and challenging developments based on the stabilization principle of the inverted pendulum.**

## I. INTRODUCTION

Control theory is a field rich in opportunities and new directions [1] dealing with disciplines and methods that leads to an automatic decision process in order to improve the performance of a control system. The evolution of control theory is related to research advances on technology, theoretical controller design methods and their real-time implementation [2, 3]. It is important to note that evolution of control theory is also closely related to education [4, 5]. The next generations of control students must receive the scientific and pedagogical supports required to verify conventional techniques, develop new tools and techniques and verify their realization [6]. If the student has no opportunity to bring into the light with realistic control problems subjected to the difficulties, training has failed [7]. The main recommendation is then to integrate into the research laboratories and curriculum education experimental projects to bridge the gap between the acquirement of scientific knowledge and the solution of practical problems.

In recent Years, projects on the themes of robotics [8, 9] and mechatronics [8, 10, 11, 12] are the most attractive for students. In this framework, many interesting robotic benchmark systems exist in the literature. The inverted pendulum system was always considered, among others, the most fundamental benchmark. Different versions of this system exist offering a variety of interesting control challenges.

This paper, proposes to enhance the wealth of this benchmark and attempt to provide an overall picture of historical, current and trend developments in control theory and robotics based on its simple structure. The paper is organized as follows: The most common robotic benchmarks and the different versions of the inverted pendulum will be exposed in the next section. The wealth of the inverted pendulum model in education will be pointed in section 3. In section 4, few real and virtual experiences will be exposed. Different control design techniques will be surveyed through application to this benchmark in section 5. Trendy robotic problems based in the inverted pendulum stabilization principle will be finally raised in section 6.

## II. ROBOTIC BENCHMARKS: AN OVERVIEW

Many robotic benchmark systems of high interest exist in the literature and frequently used for teaching and research in control theory. They are typically used to realize experimental models, validate the efficiency of emerging control techniques and verify their implementation. The most common robotic benchmarks are the Acrobot [13], the Pendubot [14], the Furuta Pendulum (see Fig.1) [15], the inverted pendulum [16], the Reaction Wheel Pendulum (see Fig.2) [17], the bicycle [18], the VTOL aircraft [19], the Beam-and-Ball system [20] and the TORA [21].

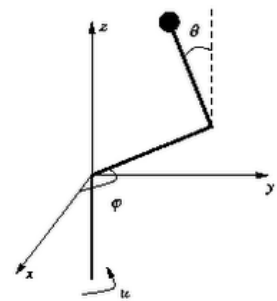

Figure 1. Furuta Pendulum

In spite of its simple structure, the inverted pendulum is considered, among the last examples, the most fundamental benchmark. Different versions of the inverted pendulum benchmark exist offering a variety of interesting control challenges.

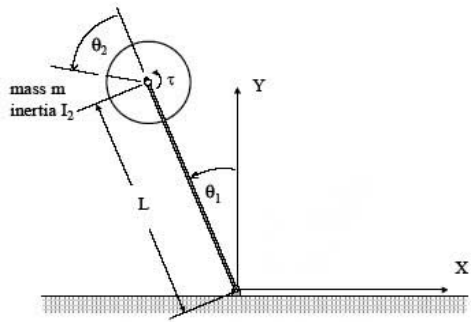

Figure 2. Inertia-Wheel Pendulum

The most familiar types are the rotational single-arm pendulum (see Fig.3) [22], the cart inverted pendulum (see Fig.4) [23], and the double inverted Pendulum [24].

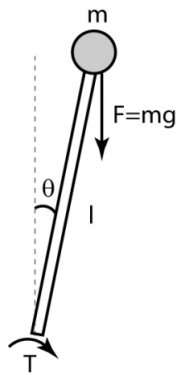

Fig.3. Rotational single-arm pendulum

The less common versions are the rotational two-link pendulum [25], the parallel type dual inverted pendulum [26], the triple inverted pendulum [27], the quadruple inverted pendulum [28] and the 3D or spherical pendulum [29].

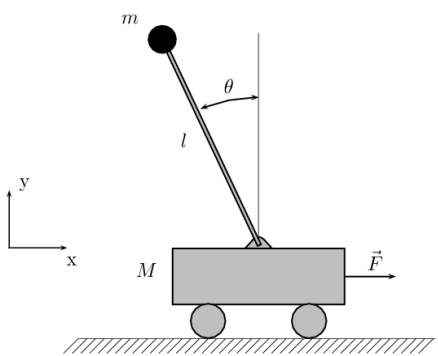

Figure 4. Cart inverted pendulum

## III. THE INVERTED PENDULUM BENCHMARK IN CONTROL THEORY EDUCATION

Since the 1950s, the inverted pendulum benchmark, especially the cart version, was used for teaching linear feedback control theory [30] to stabilize open-loop unstable systems. The first solution to this problem was described by Roberge in 1960 [31] and then by Schaefer, and Cannon in 1966 [32]. This benchmark was considered in many references as a typical root-locus analysis example [33]. Subsequently, it has been also used in many books to solve the linear optimal control problem [34] and the complex nonlinear control problem [35] for unstable systems.

In spite of the simplicity of its structure, an inverted pendulum system is a typical nonlinear dynamic system including a stable equilibrium point when the pendulum is at pending position and an unstable equilibrium point when the pendulum is at upright position. When the system is moved up from the pending position to the upright position, the model is strongly nonlinear with the pendulum angle.

The principal control task considered for the cart pendulum version is to swing up the pendulum from the stable equilibrium point to the unstable equilibrium point, and then balance the pendulum at the upright position, and further move the cart to a specified position by driving it right and left. The more general problem is guiding the pendulum from any arbitrary initial condition to the upright equilibrium and stabilizes the cart in a desired position.

## IV. REAL DEVICES, VIRTUAL MODELS AND WEB-BASED LABORATORIES

The simple structure of the inverted pendulum model allowing carrying out experimental validations is one of the key motivations of using this benchmark in education and in research. Several real devices, virtual models or web based remote control laboratories are then developed.

- **Real control laboratories**

Many real experimental models of the inverted pendulum are performed allowing to students and researches to validate the efficiency of several emerging control techniques and verify their implementation [36, 37, 38]. In many cases, common integrated software's like Matlab or LabVIEW are used to support the design of controllers, the analysis of control system and also the real-time implementation of controllers [39, 40, 41].

- **Virtual reality models**

Today, physical models might be considered old fashioned, unreliable and expensive. There are many reasons to replace them by virtual reality enabled by the computing power of computers and software tools. One may find good examples of virtual control Laboratories of pendulum-car benchmark in [42, 43, 44].

- **Web-based laboratory for remote control**

Distance education as well as remote experimentation, is a modern field that aims to deliver education to students who are not physically present on their tested instrumentation. Nowadays, many students communicate with the educational material through Internet-based technologies. Time savings, sharing resources of expensive equipment and individual access to experimentation are the main factors that motivate remote laboratories over the world. Remote pioneered projects based on the inverted pendulum can be found in [45, 46, 47, 48].

## V. THE INVERTED PENDULUM BENCHMARK IN CONTROL THEORY RESEARCH

In control theory, a rich collection of powerful and successful synthesis methods are available today for which many academic books and survey papers are devoted. In this framework, the inverted pendulum has maintained for at least fifty years its usefulness to illustrate almost all emerging ideas in this field (see Table 1).

TABLE I. SURVEY OF CONTROL DESIGN TECHNIQUES ILLUSTRATED BY THE INVERTED PENDULUM

| Control design | Academic Books | Survey papers | Research papers with illustration |
|---|---|---|---|
| Bang-Bang control | - | [49] | [50, 51] |
| Fuzzy logic control | [52] | [53, 54] | [55,56] |
| Neural Network control | [57, 58] | [59] | [60] |
| PID Adaptive control | [61] | [62, 63, 64, 65] | [66] |
| Robust control | [67, 68, 69] | [70,71] | [72] |
| Energy based control | [35,73,74,75] | [76, 77] | [22,78,79,80,81,82, 83, 84,85] |
| Hybrid control | [86, 87, 88] | [89,90,91,92, 93] | [38,94, 95,96,97] |
| Sliding mode control | [98,99,100] | [101,102] | [103,104,105] |
| Time optimal control | 106, 107, 108] | [109] | [110,111,112,113] |
| Predictive control | [114,115,116,117,118] | [118,119,120,121] | [122,123,124,125] |
| Singular perturbation | [126] | [127, 128] | [129] |
| Feedforward control | - | [130] | [131, 132] |

## VI. TRENDS IN ROBOTICS BASED ON THE INVERTED PENDULUM STABILIZATION PRINCIPLE

Based on the inverted pendulum stabilization principle, many trendy technologies in robotics will fertilize quite new control applications. The recent major accomplishments are:

- **Control of under-actuated robotic systems**

Underactuated robotic systems are systems with fewer independent control actuators than degrees of freedom to be controlled. In recent years, the need for analysis and control of underactuated robotic systems arises in many practical applications [79, 133, 134]. The cart inverted pendulum is a typical system of under-actuated robotic systems [135].

- **Design of mobile inverted pendulums**

Design and implementation of mobile wheeled inverted pendulum systems have induced a lot of attention recently [136,137,138] and at least one commercial product, the Segway, see Fig.5, is offered [139]. Such vehicles are of interest because they have a small trail.

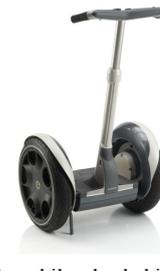

Fig.5. The Segway: A mobile wheeled inverted pendulum [139]

- **Gait pattern generation for humanoid robots**

The control of humanoid robots [140] is a challenging task in recent years due to the hard-to-stabilize dynamics. Gait pattern generation is a key problem [141, 142]. In order to simplify the trajectory generation, many studies make use of the analogy between bipedal gait and the inverted pendulum motion [143]. The Linear Inverted Pendulum Model (LIPM) [144, 145, 146, 147, 148, 149] is a trendy research topic in this framework. LIPM algorithms are generally joined to Zero Moment Point concept [150] to ensure stable gait (see Fig.6).

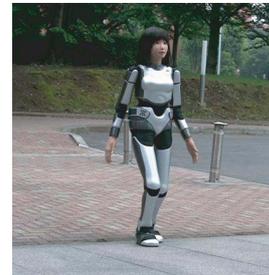

Fig.6. HRP-4C Humanoid: stable walking under linear inverted pendulum model [150]

## VII. CONCLUSION

In this paper, it has been shown that the inverted pendulum system is a fundamental benchmark in education and research in control theory. The particular interest of this application lies on its simple structure and the wealth of its model. The richness of the model has illustrated its usefulness to illustrate all emerging ideas in control theory and integrate trendy technologies and challenging applications in robotics.